\documentclass[11pt,a4paper]{article}
\usepackage[hyperref]{emnlp2018}
\usepackage{times}
\usepackage{latexsym}
\usepackage{multirow}
\usepackage{graphicx}
\usepackage{amsmath}
\usepackage{tikz-dependency}
\usepackage{url}
\usepackage[utf8]{inputenc}
\usepackage{caption,subcaption}

\usepackage{todonotes}
\usepackage{enumitem}

\aclfinalcopy 

\newcommand{\sect}[1]{Section \ref{sec:#1}}
\newcommand{\tab}[1]{Table \ref{tab:#1}}
\newcommand{\fig}[1]{Fig.\ \ref{fig:#1}}

\newcommand{\baseline}{\textsc{baseline}}
\newcommand{\combined}{\textsc{combined}}
\newcommand{\pext}{\textsc{+ext}}
\newcommand{\mext}{\textsc{$-$ext}}
\newcommand{\pchar}{\textsc{+char}}
\newcommand{\mchar}{\textsc{$-$char}}
\newcommand{\ppos}{\textsc{+pos}}
\newcommand{\mpos}{\textsc{$-$pos}}

\newcommand{\pchart}{\textsc{+ch-24}}
\newcommand{\pcharo}{\textsc{+ch-100}}
\newcommand{\pcharf}{\textsc{+ch-500}}

\newcommand{\train}{\emph{train}}
\newcommand{\dev}{\emph{dev}}

\title{An Investigation of the Interactions Between Pre-Trained Word Embeddings, Character Models and POS Tags in Dependency Parsing}

\author{Aaron Smith~~~~~~Miryam de Lhoneux~~~~~~Sara Stymne~~~~~~Joakim Nivre\\[1mm]
Department of Linguistics and Philology, Uppsala University}

\begin{document}
\maketitle
\begin{abstract}
We provide a comprehensive analysis of the interactions between pre-trained word embeddings, character models and POS tags in a transition-based dependency parser.
While previous studies have shown POS information to be less important in the presence of character models, we show that in fact there are complex interactions between all three techniques.
In isolation each produces large improvements over a baseline system using randomly initialised word embeddings only, but combining them quickly leads to diminishing returns.
We categorise words by frequency, POS tag and language in order to systematically investigate how each of the techniques affects parsing quality.
For many word categories, applying any two of the three techniques is almost as good as the full combined system. Character models tend to be more important for low-frequency open-class words, especially in morphologically rich languages, while POS tags can help disambiguate high-frequency function words.
We also show that large character embedding sizes help even for languages with small character sets, especially in morphologically rich languages.
\end{abstract}

\section{Introduction}
The last few years of research in natural language processing (NLP) have witnessed an explosion in the application of neural networks and word embeddings.
In tasks ranging from POS tagging to reading comprehension to machine translation, a unique dense vector is learned for each word type in the training data.
These word embeddings have been shown to capture essential semantic and morphological relationships between words \citep{mikolov2013distributed}, and have precipitated the enormous success of neural network-based architectures across a wide variety of NLP tasks \citep{plank2016multilingual,dhingra2016gated,vaswani2017attention}.

When task-specific training data is scarce or the morphological complexity of a language leads to sparsity at the word-type level, word embeddings often need to be augmented with sub-word or part-of-speech (POS) tag information in order to release their full power \citep{Kim2015charLM,sennrich2016BPE,chen2014fast}.
Initialising vectors with embeddings trained for a different task, typically language modelling, on huge unlabelled corpora has also been shown to improve results significantly \citep{dhingra2017comparative}.
In dependency parsing, the use of character \citep{Ballesteros2015charParsing} and POS \citep{dyer2015transition} models is widespread, and the majority of parsers make use of pre-trained word embeddings \citep{zeman-EtAl:2017:K17-3}.

While previous research has examined in detail the benefits of character and POS models in dependency parsing and their interactions \citep{Ballesteros2015charParsing,dozat2017stanford}, there has been no systematic investigation into the way these techniques combine with the use of pre-trained embeddings.
Our results suggest a large amount of redundancy between all three techniques: in isolation, each gives large improvements over a simple baseline model, but these improvements are not additive.
In fact combining any two of the three methods gives similar results, close to the performance of the fully combined system.

We set out to systematically investigate the ways in which pre-trained embeddings, character and POS models contribute to improving parser quality.
We break down results along three dimensions---word frequency, POS tag, and language---in order to tease out the complex interactions between the three techniques. 
Our main findings can be summarized as follows:
\begin{itemize}[noitemsep,topsep=4pt]
\item For all techniques, improvements are largest for low-frequency and open-class words and for morphologically rich languages.
\item These improvements are largely redundant when the techniques are used together.
\item Character-based models are the most effective technique for low-frequency 
words.
\item Part-of-speech tags are potentially very effective 
for high-frequency function words, but current state-of-the-art taggers are not accurate enough to take full advantage of this.
\item Large character embeddings are helpful for morphologically rich languages, regardless of character set size.
\end{itemize}

\section{Related Work}
\label{sec:related-work}
\citet{chen2014fast} introduced POS tag embeddings: a learned dense representation of each tag designed to exploit semantic similarities between tags.
In their greedy transition-based parser, the inclusion of these POS tag embeddings improved labelled attachment score (LAS) by 1.7 on the English Penn Treebank (ETB) and almost 10 on the Chinese Penn Treebank (CTB).
They also tested the use of pre-trained word embeddings for initialisation of word vectors, finding gains of 0.7 for PTB and 1.7 for CTB.

\citet{dyer2015transition} in their Stack Long Short-Term Memory (LSTM) dependency parser, show that POS tag embeddings in their architecture improve LAS by 0.6 for English and 6.6 for Chinese.
Unlike \citet{chen2014fast}, they do not use pre-trained word embeddings for initialisation, instead concatenating them as a fixed vector representation to a separate randomly-initialised learned representation. 
This leads to improvements in LAS of 0.9 and 1.6 of English and Chinese, respectively.

Following on from the work of \citet{dyer2015transition}, \citet{Ballesteros2015charParsing} introduced the first character-based parsing model. 
They found that a model based purely on character information performed at the same level as a model using a combination of word embeddings and POS tags.
Combining character and POS models produced even better results, but they conclude that POS tags are less important for character-based parsers.
They also showed that character models are particularly effective for morphologically rich languages, but that performance remains good in languages with little morphology, and that character models help substantially with out-of-vocabulary (OOV) words, but that this does not fully explain the improvements they bring.
The use of pre-trained embeddings was not considered in their work.

\citet{kiperwasser16}, in the transition-based version of their parser based on BiLSTM feature extractors, found that POS tags improved performance by 0.3 LAS for English and 4.4 LAS for Chinese.
Like \citet{dyer2015transition}, they concatenate a randomly-initialised word embeddings to a pre-trained word vector; however in this case the pre-trained vector is also updated during training.
They find that this helps LAS by 0.5--0.7 for English and 0.9--1.2 for Chinese, depending on the specific architecture of their system.

\citet{dozat2017stanford}, building on the graph-based version of \citet{kiperwasser16}, confirmed the relationship between character models and morphological complexity, both for POS tagging and parsing.
They also examined the importance of the quality of POS tags on parsing, showing that their own tagger led to better parsing results than a baseline provided by UDPipe v1.1 \citep{straka2016udpipe}.



\section{The Parser}
\label{sec:parser}
We use and extend UUParser\footnote{\url{https://github.com/UppsalaNLP/uuparser}} \citep{delhoneux-EtAl:2017:K17-3,smith2018st}, a variation of the transition-based parser of \citet{kiperwasser16} (K\&G).
The K\&G architecture can be adapted to both transition- and graph-based dependency parsing, and has quickly become a \emph{de facto} standard in the field \citep{zeman-EtAl:2017:K17-3}.
In a K\&G parser, BiLSTMs \citep{hochreiter1997long,graves2008bilstms} are employed to learn useful representations of tokens in context.
A multi-layer perceptron (MLP) is trained to predict transitions and possible arc labels, taking as input the BiLSTM vectors of a few tokens at a time.
Crucially, the BiLSTMs and MLP are trained together, enabling the parser to learn very effective token representations for parsing.
For further details we refer the reader to \citet{nivre2008algorithms} and \citet{kiperwasser16}, for transition-based parsing and BiLSTM feature extractors, respectively.

Our version of the K\&G parser is extended with a \textsc{Swap} transition to facilitate the construction of non-projective dependency trees \cite{nivre09acl}. We use a static-dynamic oracle to allow the parser to learn from non-optimal configurations at training time in order to recover better from mistakes at test time, as described in \citet{delhoneux17arc}. 

In this paper we experiment with a total of eight variations of the parser, where the difference between each version resides in the vector representations $x_i$ of word types $w_i$ before they are passed to the BiLSTM feature extractors (see Section 3 of \citet{kiperwasser16}).
%
%
%
In the simplest case, we set $x_i$ equal to the word embedding $e^r(w_i)$:
\begin{equation*}
  x_i = e^r(w_i)
\end{equation*}
The superscript $r$ refers to the fact that the word embeddings are initialised randomly at training time. 
This is the setup in our \baseline{} system.

For our \pchar{} system, the word embedding $e^r(w_i)$ is concatenated to a character-based vector, obtained by running a BiLSTM over the characters $ch_{1:m}$ of $w_i$:
\begin{equation*}
    x_i = e^r(w_i) \circ \text{BiLSTM}(ch_{1:m})
\end{equation*}    

In the \ppos{} setting, the word embedding is instead concatenated to an embedding $p(w_i)$ of the word's universal POS tag \citep{nivre+16UD}:
\begin{equation*}
    x_i = e^r(w_i) \circ p(w_i)
\end{equation*} 
This scenario necessitates knowledge of the POS tag of $w_i$; at test time, we therefore need a POS tagger to provide predicted tags.

In another version of our parser (\pext{}), pre-trained embeddings are used to initialise the word embeddings.\footnote{This strategy proved more successful in preliminary experiments than others for incorporating pre-trained embeddings discussed in \sect{related-work}.} We use the superscript $t$ to distinguish these from randomly initialised vectors:
\begin{equation*}
  x_i = e^t(w_i)
\end{equation*}
We use the embeddings that were released as part of the 2017 CoNLL Shared Task on Universal Dependency Parsing (CoNLL-ST-17) \citep{zeman-EtAl:2017:K17-3}.
Words in the training data that do not have pre-trained embeddings are initialised randomly.
At test time, we look up the updated embeddings for all words seen in the training data; OOV words are assigned their un-updated pre-trained embedding where it exists, otherwise a learnt OOV vector.


In our \combined{} setup, we include pre-trained embeddings along with the character vector and POS tag embedding:
\begin{equation*}
\label{eq:combined}
    x_i = e^t(w_i) \circ \text{BiLSTM}(ch_{1:m}) \circ p(w_i)
\end{equation*} 
The three remaining versions of the vector $x_i$ constitute all possible combinations of two techniques of pre-trained embeddings, the character model and POS tags.
%
%
We refer to these versions of the parser as \mext{}, \mchar{}, and \mpos{}, respectively.


\section{Experimental setup}
\label{sec:setup}
%

\subsection{Data}
\label{sec:data}
We ran our experiments on nine treebanks from Universal Dependencies \citep{nivre+16UD} (v2.0): Ancient Greek PROIEL, Arabic, Chinese, English, Finnish, Hebrew, Korean, Russian and Swedish.
Inspired partially by \citet{de2017old}, these treebanks were chosen to reflect a diversity of writing systems, character set sizes, and morphological complexity.
As error analysis is carried out on the results, we perform all experiments on the \dev{} data sets.

\tab{treebank-sizes} shows some statistics of each treebank. 
Of particular note are the large character set sizes in Chinese and Korean, an order of magnitude bigger than those of all other treebanks. 
The high type-token ratio for Finnish, Russian and Korean also stands out; this is likely due to the high morphological complexity of these languages.

%
\begin{table}[!h]
\vspace{2mm}
\begin{center}
\begin{tabular}{ | l | r | r | r | r | }
 \hline
 Treebank & \multicolumn{2}{|c|}{Sentences} & TTR & Chars \\
 \hline
 Ancient Greek & 14864 & 1019 & 0.15 & 179 \\
 Arabic & 6075 & 909 & 0.10 & 105 \\ 
 Chinese & 3997 & 500 & 0.16 & 3571 \\
 English & 12534 & 2002 & 0.07 & 108 \\  
 Finnish & 12217 & 1364 & 0.26 & 244 \\
 Hebrew & 5241 & 484 & 0.11 & 53 \\
 Korean & 4400 & 950 & 0.46 & 1730 \\
 Russian & 3850 & 579 & 0.30 & 189 \\
 Swedish & 4303 & 504 & 0.16 & 86 \\
 \hline
\end{tabular}
\caption{Treebank statistics. Number of sentences in \train{} and \dev{} sets, type-token ratio (TTR), and character set size. 
}
\vspace{-6mm}
\label{tab:treebank-sizes}
\end{center}
\end{table}

\subsection{Parser settings}
\label{sec:settings}
The parser is trained three times for each language with different random seeds for 30 epochs each.
At the end of each epoch we parse the \dev{} data and calculate LAS.
For each training run, results are averaged over the five best epochs for each language.
In this way, we attempt to make our results more robust to variance due to randomness in the training procedure.\footnote{Changing the random seed has been shown to produce results that appear statistically significant different in neural systems \citep{reimers2017reporting}.}
Our macro-averaged scores are based on a total of 135 different epochs (3 random seeds $\times$ 5 best epochs $\times$ 9 languages).

\tab{hyperparameters} shows the embedding sizes we  found to produce best results in preliminary work and which we use in all experiments in \sect{results}.
\begin{table}[t]
\begin{center}
\begin{tabular}{  l | c  }
 \hline
  Word embedding size & 100 \\
  \hline
  Character embedding size & 500 \\
  Character BiLSTM output size & 100 \\
  \hline
  POS tag embedding size & 20 \\
  \hline
\end{tabular}
\caption{Embedding sizes.}
\vspace{-6mm}
\label{tab:hyperparameters}
\end{center}
\end{table}
Note our unusually large character embedding size; we will discuss this in more detail in \sect{chars}.
We use predicted UPOS tags from the system of \citet{dozat2017stanford} for experiments with POS tags,\footnote{Available at \newline \url{https://web.stanford.edu/~tdozat/}.} other than in \sect{tagger} where we compare results with different taggers and gold POS tags, in order to set a ceiling on the potential gains from a perfect POS tagger. 
For all other hyperparameters we use default values \citep{smith2018st}. 

\subsection{Analysis}
\label{sec:analysis}
The hypothesis underlying our choice of analysis is that the three techniques under study here---pre-trained embeddings, character vectors and POS tag embeddings---affect words differently depending on their frequencies, POS tags, and the language of the sentence.
We do not claim this to be an exhaustive list; many other dimensions of analysis are clearly possible (dependency relation would be another obvious choice for example), but we believe that these are likely to be three of the most informative factors.
In the frequency and POS tag cases, we want to examine the overall contribution to LAS of words from each category.
We expect changing the representation of a token to affect how likely it is to be assigned the correct head in the dependency tree, but also how likely it is to be assigned correctly as the head of other words.
We thus introduce a new metric for this part of the analysis: the head and dependents labelled attachment score, which we refer to as HDLAS.

When calculating HDLAS, the dependency analysis for a given token is only considered correct if the token has the correct labelled head \emph{and} the complete set of correctly labelled dependents.
This is a harsher metric than LAS, which only considers whether a token has the correct labelled head.
Note that when calculating HDLAS for all tokens in a sentence, each dependency relation is counted twice, once for the head word and once for the dependent.
It only makes sense to use this metric when analysing individual tokens in a sentence, or when grouping tokens into different categories across multiple sentences.
%

\subsubsection{Frequency}
\label{sec:frequency}
In this analysis, we first label each token in the \dev{} data for each language by its relative frequency in the \train{} data, with add-one smoothing.\footnote{
The smoothing ensures that OOV tokens, those that appear in \dev{} but not \train{}, are not assigned zero frequency; this alleviates the problem of taking log(0) in the subsequent conversion to log relative frequency.} 
Frequency categories are created by rounding the log relative frequency down to the nearest integer.
We calculate the HDLAS for each frequency category for each language, before macro-averaging the results across the nine languages to produce a final score for each frequency class.

\subsubsection{POS tag}
\label{sec:pos-analysis}
In this case, we label each word from the \dev{} data by its gold POS tag, before calculating HDLAS for each category and taking the macro average across languages.
Here the total number of tokens in each category varies across several orders of magnitude: the most common category NOUNs make up 26.0\% of all words, while the smallest class SYM represents just 0.1\%.
For this reason, and to make our graphs more readable, we do not show results for the six smallest categories: INTJ, NUM, PART, SCONJ, SYM, and X.

\subsubsection{Language}
\label{sec:lang-analysis}
Here we consider LAS directly for each language; the HDLAS metric used in the previous two sections is not relevant as all tokens in a given sentence are assigned to the same category determined by the language of the sentence. 

\section{Results}
\label{sec:results}
\tab{results} gives the LAS for each of the eight systems described in \sect{parser}. 
\begin{table}[t]
\begin{center}
\begin{tabular}{ | l | c | l | c | }
 \hline
 \baseline{} & 67.7 & \combined{} & 81.0 \\
 \hline
 \pext{} & 76.1 & \mext{} & 79.9 \\
 \pchar{} & 78.3 & \mchar{} & 79.2 \\ 
 \ppos{} & 75.9 & \mpos{} & 80.3 \\  
 \hline
\end{tabular}
\caption{Mean LAS across nine languages for a baseline system employing randomly-initialised word embeddings only, compared to three separate systems using pre-trained word embeddings (\pext{}), a character model (\pchar), and POS tags (\ppos). 
Scores are also shown for a combined system that utilises all three techniques 
and corresponding systems where one of the three techniques is ablated (\mext{}, \mchar{} and \mpos{}).}
\vspace{-6mm}
\label{tab:results}
\end{center}
\end{table}
We observe that pre-trained embeddings (+8.4), the character model (+10.6) and POS tags (+8.2) all give large improvements in LAS over the baseline system. 
The combined system is the best overall, but the improvement of 13.3 LAS is far from the sum of its components.
Employing two of the techniques at a time reduces LAS by only 0.7--1.8 compared to the combined system.

\subsection{Frequency}
\label{sec:analysis-freq}
\begin{figure}[b]
\centering
\includegraphics[width=\linewidth]{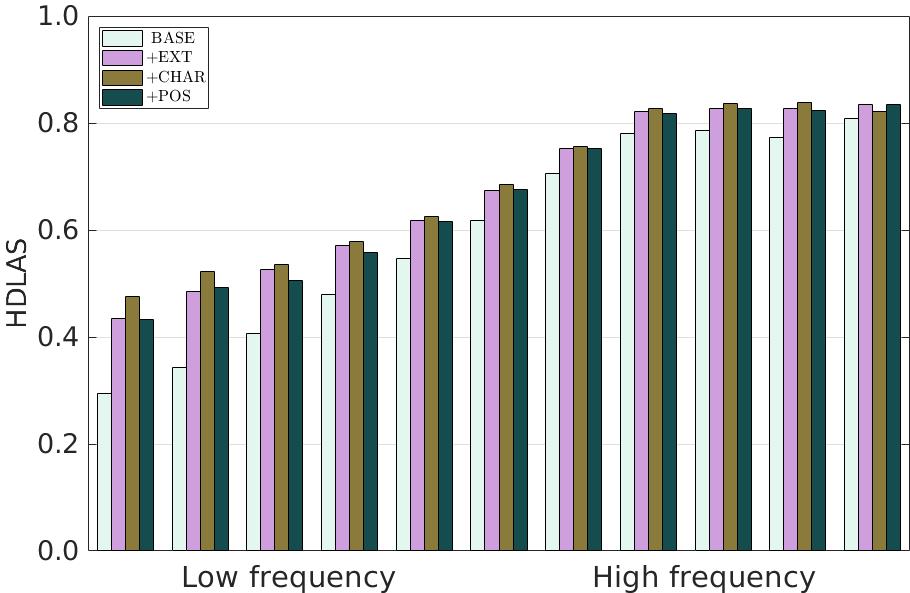}
\caption{\label{fig:baseline+all+combined-freq}\baseline{} system compared to pre-trained embeddings (\pext{}), character model (\pchar{}) and POS tags (\ppos{}).}
\end{figure}

\begin{figure}[t]
\centering
\includegraphics[width=\linewidth]{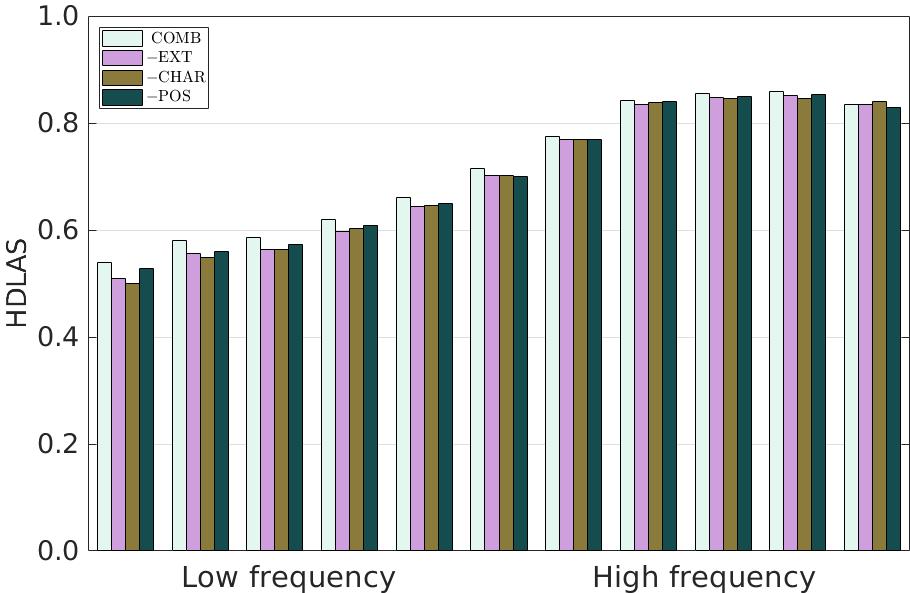}
\caption{\label{fig:baseline-all+combined-freq}\combined{} system compared to ablated systems where pre-trained embeddings (\mext{}), character models (\mchar{}) and POS tags (\mpos{}) are removed.}
\vspace{-4mm}
\end{figure}

\fig{baseline+all+combined-freq} and \fig{baseline-all+combined-freq} compare systems by word frequency.
%
As expected, accuracy improves with frequency for all systems: the parser does better with words it sees more often during training.
There is a levelling off for the highest frequency words, probably due to the fact that these categories contain a small number of highly polysemous word types.

\fig{baseline+all+combined-freq} demonstrates a clear trend in the improvement achieved by each of the individual techniques over the baseline, with larger gains for lower frequency words.
This confirms a result from \citet{Ballesteros2015charParsing}, who found that character models help substantially with OOV words.
We can generalise this to say that character models improve parsing quality most for low frequency words (including OOV words), and that this is also true, albeit to a slightly lesser effect, of POS tags and pre-trained word embeddings.
It is notable however that HDLAS increases universally across all frequency classes: even the highest frequency words benefit from enhancements to the basic word representation.

%
%
%
What immediately stands out in \fig{baseline-all+combined-freq} is that for mid- and high frequency words, there is little difference in HDLAS between different combinations of two of the three techniques, and for the highest frequency words this is at a level almost indistinguishable from the full \combined{} system.
The slight improvements we see for \combined{} in \tab{results} compared to the three ablated systems thus principally also come from the low-frequency range. 

\subsection{POS tags}
\label{sec:analysis-pos}
\begin{figure}[!h]
\vspace{2mm}
\centering
\includegraphics[width=\linewidth]{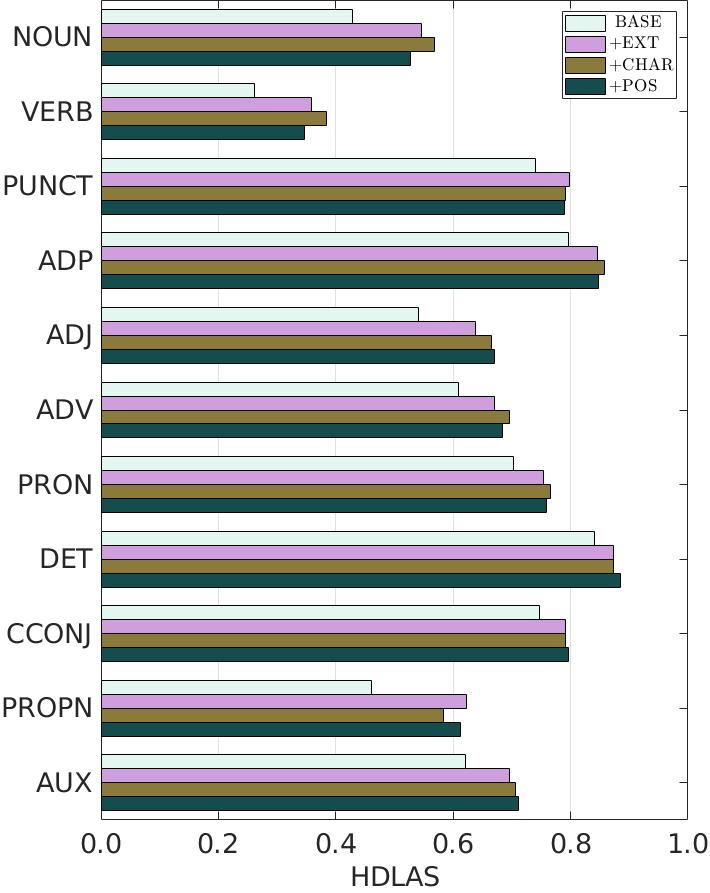}
\caption{Comparison by POS tag of \baseline{} system to \pext{}, \pchar{}, and \ppos{}. Tags are sorted by frequency.}
\vspace{-4mm}
\label{fig:baseline+all+combined-pos}
\end{figure}

In \fig{baseline+all+combined-pos} systems are compared by POS tag.
We observe a universal improvement across all POS tags for each of the three variations of the system compared to the baseline.
However, it is notable that the biggest gains in HDLAS are for open word classes: NOUNs, VERBs and ADJs.
As these make up a large overall proportion of words, these differences have an overall relatively large impact on LAS.

For the most frequent POS categories NOUN and VERB we again see a clear victory for the character model (note that while these POS categories are frequent, they contain a large number of low-frequency words).
Overall the character model succeeds best for the open-class POS categories, while having the right POS tag is marginally better for closed-class categories such as DET, CCONJ, and AUX.
It is interesting that the character model is not as strong for PROPN, despite the fact that these are open-class low-frequency words; for these words pre-trained embeddings are the best single technique.
This may be due to the fact that the rules governing the composition of names at the character level are different from other words in the language.

It is perhaps surprising that the advantage of POS tag embeddings is not greater when it comes to auxiliary verbs, for example, where the distinction from main verbs can be difficult and crucial for a correct syntactic analysis.
The reason probably lies in the fact that this distinction is equally difficult for the POS tagger.
We will investigate this further in \sect{tagger}.

%
%

%
%


\subsection{Language}
\label{sec:analysis-lang}
\begin{figure}[!t]
\centering
\includegraphics[width=\linewidth]{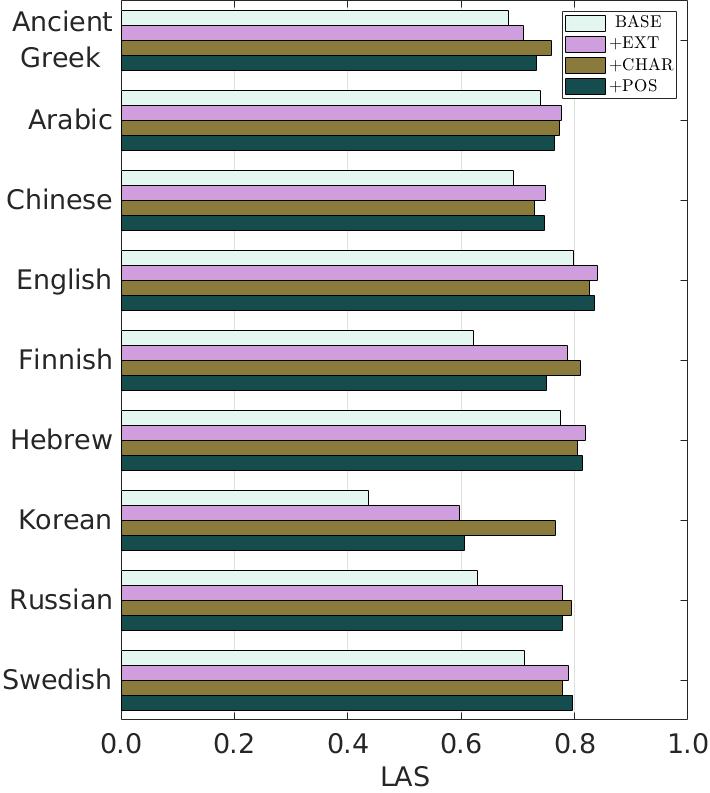}
\caption{Comparison by language of \baseline{} system to \pext{}, \pchar{}, and \ppos{}.}
\vspace{-5mm}
\label{fig:baseline+all+combined-lang}
\end{figure}

\fig{baseline+all+combined-lang} compares the systems by language.
Once again improvement is universal for each system compared to the baseline.
There are however substantial differences between languages.
The three biggest overall improvements are for Finnish, Korean and Russian, with a particularly notable increase in the Korean case.
This suggests that the baseline model struggles to learn adequate representations for each word type in these languages.
These are the three languages we identified in \sect{data} as having high type-token ratios in their training data.
It is also notable that the character model becomes more important compared to other methods for these three languages.
In fact, despite the overall superiority of the character model (see \tab{results}), it is only the best single technique for 4 of the 9 languages, the three already mentioned plus Ancient Greek.

%
%

\section{Character Embedding Size}
\label{sec:chars}
All results with character models observed thus far make use of a character embedding of dimension 500.
This value is large compared to typical sizes used for character models \citep{Kim2015charLM,Ballesteros2015charParsing}.
A common belief is that larger character embedding sizes are justified for languages with larger character set sizes such as Chinese: in other words, the embedding size should be related to the number of entities being embedded \citep{shao2018segmenting}.

In \tab{char-results}, we show how LAS varies with a few values of this hyperparameter when averaged across our nine-language sample.
\begin{table}[t]
\begin{center}
\begin{tabular}{ | l | c | l | c | }
 \hline
 \baseline{} & 67.7 & \mchar & 79.2 \\
 \hline
 \pchart & 76.8 & \pchart & 80.5 \\
 \pcharo & 77.7 & \pcharo & 80.6 \\ 
 \pcharf & 78.3 & \pcharf & 81.0 \\  
 \hline
\end{tabular}
\caption{Mean LAS across nine languages for \baseline{} system compared to systems with character vectors of different sizes. Comparison also shown for systems employing pre-trained word vectors and POS tag embeddings.}
\vspace{-4mm}
\label{tab:char-results}
\end{center}
\end{table}
We see a steady improvement in LAS as the character embedding size increases, both when compared to a baseline with randomly initialised word embeddings only and when compared to a system that also employs pre-trained word vectors and POS tag embeddings.\footnote{Note that for character embeddings of dimension 24, we use an output size for the character BiLSTM of 50, for character embeddings of dimension 100, we use an output size of 75, and for character embeddings of dimension 500, we use an output size of 100.
We checked in separate experiments that the improvements are not simply due to the increase in output size.}

It is particularly interesting to break down the effects here by language.
In \tab{chars-lang} we show results for Chinese, Finnish, Korean and Russian.
It is particularly striking that the larger character embeddings do not help for Chinese; the score for the largest character embedding size is actually marginally lower than a baseline without a character model at all.
This is despite the fact that a small character embedding improves LAS, albeit marginally, suggesting that there is some useful information in the characters even when pre-trained embeddings and POS tags are present.
Conversely, the large character models are very effective for Finnish, a treebank with a character set less than a tenth of the size of Chinese (see \tab{treebank-sizes}).

%
%
%
\begin{table}[!h]
\vspace{2mm}
\begin{center}
\scalebox{0.87}{\begin{tabular}{ | l | c | c | c | c | }
 \hline
~ & \mchar{} & \pchart{} & \pcharo{} & \pcharf{} \\
 \hline
 Chinese & 76.0 & 76.1 & 75.9 & 75.8 \\
 Finnish & 81.9 & 83.7 & 83.8 & 84.7 \\ 
 Korean & 70.1 & 78.0 & 78.2 & 79.4 \\
 Russian & 82.0 & 81.4 & 81.5 & 82.5 \\
 \hline
\end{tabular}}
\caption{Comparison by language of different character embedding sizes.}
\vspace{-2mm}
\label{tab:chars-lang}
\end{center}
\end{table}

We claim therefore that character set size is not in fact a good metric to use in determining character embedding sizes.
Our tentative explanation is that while languages like Finnish have relatively small character sets, those characters interact with each other in much more complex ways, thus requiring larger embeddings to store all the necessary information.
While there are many characters in Chinese, the entropy in the interactions between characters appears to be smaller, enabling smaller character embeddings to do just as good a job.

It is also worth noting from Tables \ref{tab:char-results} and \ref{tab:chars-lang} that, in the presence of POS tags and pre-trained embeddings, the improvement gained from increasing the character embedding size from 24 to 100 is small (0.1 LAS for Finnish, 0.2 for Korean, 0.1 for Russian; 0.1 on average across the nine treebanks).
This perhaps gives the impression of diminishing returns; that going even larger is likely to lead to ever smaller improvements.
This may be the reason that smaller character embeddings have generally been preferred previously.
However, we in fact observe a much \emph{greater} gain when increasing from 100 to 500 (0.9 for Finnish, 1.2 for Korean, 1.0 for Russian; 0.4 on average across the nine treebanks), suggesting that very large character embeddings are effective, and particularly useful for morphologically rich languages.

\section{POS tagger}
\label{sec:tagger}
In this section we apply our POS tag analysis to the effect of the POS tagger used to produce tags at test time.
We compare three setups: firstly using tags predicted by UDPipe \citep{udpipe:2017}, which was the baseline model for CoNLL-ST-2017, secondly using tags predicted by the winning Stanford system \citep{dozat2017stanford}, and thirdly using gold tags.
Note that for the Stanford system, we train on gold tags and use predicted tags at test time, while for UDPipe we train on a jackknifed version of the train data with predicted tags that was released as part of CoNLL-ST-2017.

\begin{table}[!h]
\vspace{2mm}
\begin{center}
\begin{tabular}{ | l | c | l | c | }
 \hline
 \baseline{} & 67.7 & \mpos{} & 80.3 \\
 \hline
 UDPipe & 73.4 & UDPipe & 80.2 \\
 Stanford & 75.9 & Stanford & 81.0 \\ 
 Gold & 78.4 & Gold & 83.8 \\  
 \hline
\end{tabular}
\caption{Mean LAS across nine languages for \baseline{} system compared to systems with POS tags predicted by different systems. Comparison also shown for systems employing pre-trained word vectors and a character vector.}
\vspace{-2mm}
\label{tab:pos-results}
\end{center}
\end{table}

\tab{pos-results} shows how LAS varies with the different POS taggers when averaged across the nine-language sample.
We see a clear improvement from UDPipe to Stanford and then from Stanford to gold tags over the baseline system.
This partially confirms results from \citet{dozat2017stanford}, where the Stanford tagger was found to improve parsing results significantly over the UDPipe baseline.
More surprising perhaps is the result when comparing to the \mpos{} system, which also makes use of pre-trained word embeddings and a character model.
Here, results do not improve at all by adding predicted tags from UDPipe.
Stanford tags do give an improvement of 0.7 LAS over \mpos{}, but this is a long way from the improvement of 8.2 LAS we see when adding them on top of \baseline{}.
Gold tags do however still give a big improvement over \mpos{} (3.5 LAS), suggesting strongly that both UDPipe and Stanford struggle with the decisions that would be most beneficial to parsing accuracy.

\begin{figure}[t]
\centering
\includegraphics[width=\linewidth]{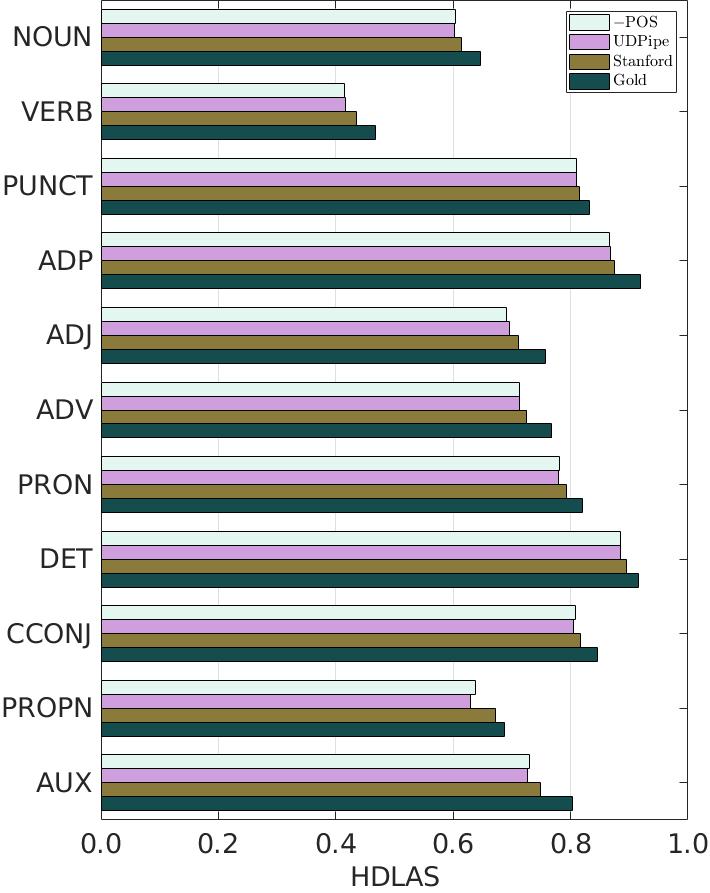}
\caption{Comparison by POS tag of POS taggers.}
\vspace{-4mm}
\label{fig:pos-pos}
\end{figure}

In \fig{pos-pos} we present the parsing results broken down by POS tag for the various POS taggers.
It is particularly notable that results when tagging with UDPipe are no better than for \mpos{}, which does not use POS tags at all, across most categories, and particularly for the closed-classes ADP, PRON, DET, CCONJ and AUX.
Stanford tags do marginally better, but access to gold tags is particularly important in these cases; we see a particularly striking improvement when ADPs and AUXs are correctly tagged over an already strong baseline.

\section{Parser speed}
\label{sec:speed}
It should be noted that increasing the character embedding size and character BiLSTM output dimension as in \sect{chars} slows down the parser during training and at test time.
We found no noticeable difference in speed between the baseline system and versions of the parser with smaller character embedding sizes (24/100), with approximately 20 sentences per second being processed on average during training and 65 sentences per second parsed at test time on the Taito super cluster.\footnote{\url{https://research.csc.fi/taito-supercluster}}
There was however a discernible difference when the character embedding size was increased to 500, with only 12 sentences processed per second during training and 44 during testing.

Adding a POS tag embedding makes no appreciable difference to parser speed,\footnote{Note that the POS tag embedding we use is small relative to the other components of the word type representation (see \tab{hyperparameters}).} but necessitates a pipeline system that first predicts POS tags (assuming gold tags are unavailable). 
The application of pre-trained embeddings, meanwhile, requires expensive pre-training on large unlabelled corpora.
Loading these embeddings into the parser takes time and can occupy large amounts of memory, but does not directly impact the time it takes to process a sentence during training or parsing.

\section{Conclusions and Future Work}
\label{sec:conclusions}
In this article we examined the complex interactions between pre-trained word vectors, character models and POS tags in neural transition-based dependency parsing.
While previous work had shown that POS tags are not as important in the presence of character models, we extend that conclusion to say that in the presence of two of the three techniques, the third is never as important.
The best system, however, is always a combination of all three techniques.

We introduced the HDLAS metric to capture the overall effect on parsing quality of changes to the representation of a particular word.
We found that all three techniques produce substantial improvements across a range of frequency classes, POS tags, and languages, but the biggest improvements for all techniques were for low-frequency, open-class words.
We suggest that this goes some way to explaining the redundancy between the three techniques: they target the same weaknesses in the baseline word-type level embedding.

We confirmed a previous result that the character model is particularly important for morphologically rich languages with high type-token ratios, and went on to show that these languages also benefit from larger character embedding sizes, whereas morphologically simpler languages make do with small character embeddings, even if the character set size is large.

POS tag embeddings can improve results for difficult closed-class categories, but our current best POS taggers are not capable of making the distinctions necessary to really take advantage of this.
The strength of pre-trained embeddings is that they are trained on much larger corpora than the task-specific data; the use of character models and POS tag embeddings however seems to allow us to generalise much better from smaller data sets, as each character and each POS tag is normally seen many times, even if each word type is rare.

We saw that increasing the character embedding size slows the parser down; whether this trade-off is worthwhile will depend on the application in question.
If accuracy is all that matters, we recommend using a fully combined system with large character embeddings in tandem with POS tags and pre-trained embeddings.
Where speed is more important, it may be worth considering a system that employs a smaller character embedding and does without POS tags, using just pre-trained embeddings.

In future work it would be interesting to investigate whether the patterns observed here also hold true for other types of models in dependency parsing; possible variations to examine include alternative character models such as convolutional neural networks, joint tagging-parsing models, and graph-based parsers.

\section*{Acknowledgments}

We acknowledge the computational resources provided by CSC in Helsinki and Sigma2 in Oslo through NeIC-NLPL (www.nlpl.eu). 
Aaron Smith was supported by the Swedish Research Council.
We would like to thank Sujoung Baeck for a valuable discussion regarding Korean morphology.

\bibliography{confs-long,refs-basic}
\bibliographystyle{acl_natbib}

\end{document}